\documentclass{article} 
\usepackage{iclr2022_conference,times}
\usepackage{paralist}

\usepackage{amsmath,amsfonts,bm}

\def\eqref#1{equation~\ref{#1}}

\def\1{\bm{1}}

\DeclareMathAlphabet{\mathsfit}{\encodingdefault}{\sfdefault}{m}{sl}
\SetMathAlphabet{\mathsfit}{bold}{\encodingdefault}{\sfdefault}{bx}{n}

\usepackage{subfig}

\usepackage{hyperref}
\usepackage{url}
\usepackage{graphicx}

\title{MVMTnet: A Multi-variate Multi-modal Transformer for Multi-class Classification of Cardiac Irregularities Using ECG Waveforms and Clinical Notes}

\author
{Ankur Samanta \textsuperscript{1 2 3 4}
\And
Mark Karlov \textsuperscript{1 4}
\And
Meghna Ravikumar \textsuperscript{1 4}
\And
Christian McIntosh Clarke \textsuperscript{1 4}
\And
Jayakumar Rajadas \textsuperscript{2 }
\And
Kaveh Hassani \textsuperscript{1}
}

\iclrfinalcopy 
\begin{document}

\maketitle
\begin{abstract}

Deep learning provides an excellent avenue for optimizing diagnosis and patient monitoring for clinical-based applications, which can critically enhance the response time to the onset of various conditions. For cardiovascular disease, one such condition where the rising number of patients increasingly outweighs the availability of medical resources in different parts of the world, a core challenge is the automated classification of various cardiac abnormalities. Existing deep learning approaches have largely been limited to detecting the existence of an irregularity, as in binary classification, which has been achieved using networks such as CNNs and RNN/LSTMs. The next step is to accurately perform multi-class classification and determine the specific condition(s) from the inherently noisy multi-variate waveform, which is a difficult task that could benefit from (1) a more powerful sequential network, and (2) the integration of clinical notes, which provide valuable semantic and clinical context from human doctors. Recently, Transformers have emerged as the state-of-the-art architecture for forecasting and prediction using time-series data, with their multi-headed attention mechanism, and ability to process whole sequences and learn both long and short-range dependencies. The proposed novel multi-modal Transformer architecture would be able to accurately perform this task while demonstrating the cross-domain effectiveness of Transformers, establishing a method for incorporating multiple data modalities within a Transformer for classification tasks, and laying the groundwork for automating real-time patient condition monitoring in clinical and ER settings.
\end{abstract}

Keywords: Transformer, Multi-class, Multi-modal,  Cardiac abnormality classification, Transfer learning, ECG waveforms, Clinical notes, MVMTnet\footnote{\href{https://bit.ly/MVMTnet-Transformer_ECGClassification}{Codebase:} \url{bit.ly/MVMTnet-Transformer_ECGClassification} \\ 
\texttt{1 University of Toronto: 27 King's College Cir, Toronto, ON M5S, Canada} \\ 
\texttt{2 Advanced Drug Delivery and Regenerative Biomaterials Laboratory, Stanford Cardiovascular Institute - Pulmonary and Critical Care: 1050 Arastradero Rd, Palo Alto, CA 94304} \\ 
\texttt{3 Corresponding author contact: ankur.samanta@mail.utoronto.ca} \\ 
\texttt{4 Equal contributions}}

\section{Introduction}
\label{Introduction}

Deep learning has revolutionized medical signal processing and is able to outperform traditional Electrocardiogram (ECG) analysis for cardiac diagnostics \citep{Smith2019}. ECG waveforms, which depict the electrical activity of the heart, are a time series representation of the heart's voltage. Different ECG machines can have multiple leads, each representing different directions of cardiac activation \citep{park2022}. Additionally, deep learning is shown to be the most effective strategy for analyzing clinical notes, which are unstructured text annotations that contain contextual information about the patient, such as diagnoses, symptoms, and treatments, for efficient patient diagnosis \citep{Chandru2022}. Automating the analysis of ECG signals and clinical notes is both cost-effective and allows for earlier identification of cardiovascular diseases, streamlining the diagnostic process at hospitals and enabling faster response times to changes in cardiac conditions.

Prior works tackling similar problems have used a variety of legacy networks, such as Convolutional Neural Networks (CNN) \citep{ebrahimi2020}, which have the disadvantage of losing positional and sequential features of the data that are critical for analyzing sequential data such as time-series waveforms. On the other hand, approaches using sequential models such as Recurrent Neural Networks (RNN) and Long Short-Term Memory networks (LSTM) have the disadvantage of only being able to process a single token at a time with hidden states for memory, capturing limited historic representations, being bound by a particular sequence length, and being recursive and non-parallelizable, meaning relatively inefficient and require longer training times \citep{Sherstinsky2018}. In contrast, Transformers have proven to be highly effective at analyzing sequential data, being able to process entire sequences of variable lengths at once (they are length invariant and treat the sequence as a whole unit), and use positional encodings and attention to generate and prioritize relevant variable-length connections throughout the sequence \citep{Vaswani2017}.

As such, this paper proposes the use of a novel multi-variate multi-modal Transformer to combine textual data from clinical notes with time-series electrocardiogram (ECG) waveform data to perform multi-class classification of observed cardiac abnormalities, a task which has not been accomplished to this extent with such an architecture before. Our multi-variate model is compatible with any number of ECG leads and is sequence length-invariant, but for this work, we focus on 12-lead waveforms (each sample has 12 distinct waveforms, which contain more information than otherwise uni-variate single-lead ECGs). The creation of a multi-modal architecture through the addition of a correspondent data source (clinical notes, analyzed with Clinical BERT, meaning our model inherits its pre-trained text processing capabilities for clinical settings) exhibits the potential to improve classification through valuable semantic contextual information about the medical practitioner’s observations \citep{Alsentzer2019}. Ultimately, this will corroborate the automated diagnosis provided by the Transformer, which originally would have only considered historic ECG waveform data (the waveform Transformer takes inspiration from core features of the seminal Transformer architecture by \citep{Vaswani2017}, inheriting its architectural benefits for sequential data analysis). 

Thus, the MVMTnet architecture produces an overall more accurate model and lays the groundwork for developing multi-modal neural networks for clinical applications that are effectively able to harness both medical-condition monitoring devices as well as a clinician's own insights in its analysis.

\section{Background \& Related Work}
\label{Background & Related Work}

\subsection{Multi-class classification of time series} The use of artificial neural networks for the classification of time series data is not a novel idea, and a considerable amount of literature has been published on machine learning-based techniques to classify cardiac irregularities in ECG waveforms \citep{Adams2012, bizopoulos2018, chamatidis2017, isin2017, Hatami2022}. An example of an approach to ECG multi-class classification was using a pre-trained CNN structure like AlexNet, which showed that CNN classifiers yield high accuracy \citep{Eltrass2022}. To build upon this, Lui \& Chow present a CNN-LSTM stack decoding classifier that detects MI in patients with pre-existing heart conditions. While this method was able to achieve over $90\%$ performance metric, it is limited to identifying only MI against a pool of “other” CVD types \citep{Lui2018}. Likewise, the state-of-the-art for ECG waveform classification models is generally constrained to binary classification models \citep{ebrahimi2020}.

\subsection{Multi-modal neural networks} Similarly, the concept of combining multiple modes of data to aid classification has been explored and documented well in literature, specifically with a combination of CNN and RNN architectures \citep{tan2017, jo2019, ahmad2021}. 

Recent papers show that combining time series data and clinical notes is possible and effective for forecasting \citep{Khadanga2019, deznabi2021}. CNN-LSTM networks are predominantly employed for such tasks, such as when the two modalities of magnetic resonance images and clinical data were used to predict outcomes in stroke patients \citep{Hatami2022}. Similar medical applications are explored in Khadanga and Aggarwal’s paper with ICU data and clinical notes in predicting in-hospital mortality rates. Their approach shows how to interpret time series data, such as ECG waveforms, as data with time stamps and demonstrates how to tokenize and analyze data \citep{Khadanga2019}. 

\subsection{Multi-variate Transformers for time series} Transformers are not constrained by recursive sequential processing, and thus outperform RNNs, LSTMs, and other models at forecasting time-series data such as ECG waveforms \citep{Vaswani2017}. A recent paper provided the first instance of using a generalized multi-variate Transformer for time series classification on various multi-class datasets \citep{Zerveas2020}. The model provided superior results over ROCKET, XGBoost, LSTM, and DTW\_D; however, the paper does not explore the use of another modality \citep{Zerveas2020}.

The state-of-the-art analysis shows that the multi-class classification of ECG waveforms as well as the introduction of multiple modes to a multi-variate Transformer when integrated are a novel concept, and points to effective time-series classification through the combination of these ideas. Therefore, the proposed model requires the ability to classify each heart condition available in the PTB-XL dataset, combined with the use of clinical notes for increased accuracy. A multi-variate Transformer was chosen to implement this novel multi-modal multi-class framework. 

\section{Data Processing Methods}
\label{Data Processing}

The main source of data is the PTB-XL ECG dataset \citep{Wagner2020}. This is a well-known electrocardiography (ECG) dataset that is publicly available and contains $21837$ clinical 12-lead ECGs from $18885$ patients. This dataset was selected due to its size, quality of organization, and detailed annotations. Additionally, all records were validated once by a technical expert, and most were validated another time by a cardiologist \citep{Wagner2020}. However, the raw data must be pre-processed for it to be compatible with the proposed model and useful for analysis \citep{Mishra2016}.

\begin{figure}[h]
\begin{center}
\includegraphics[width=0.9\textwidth]{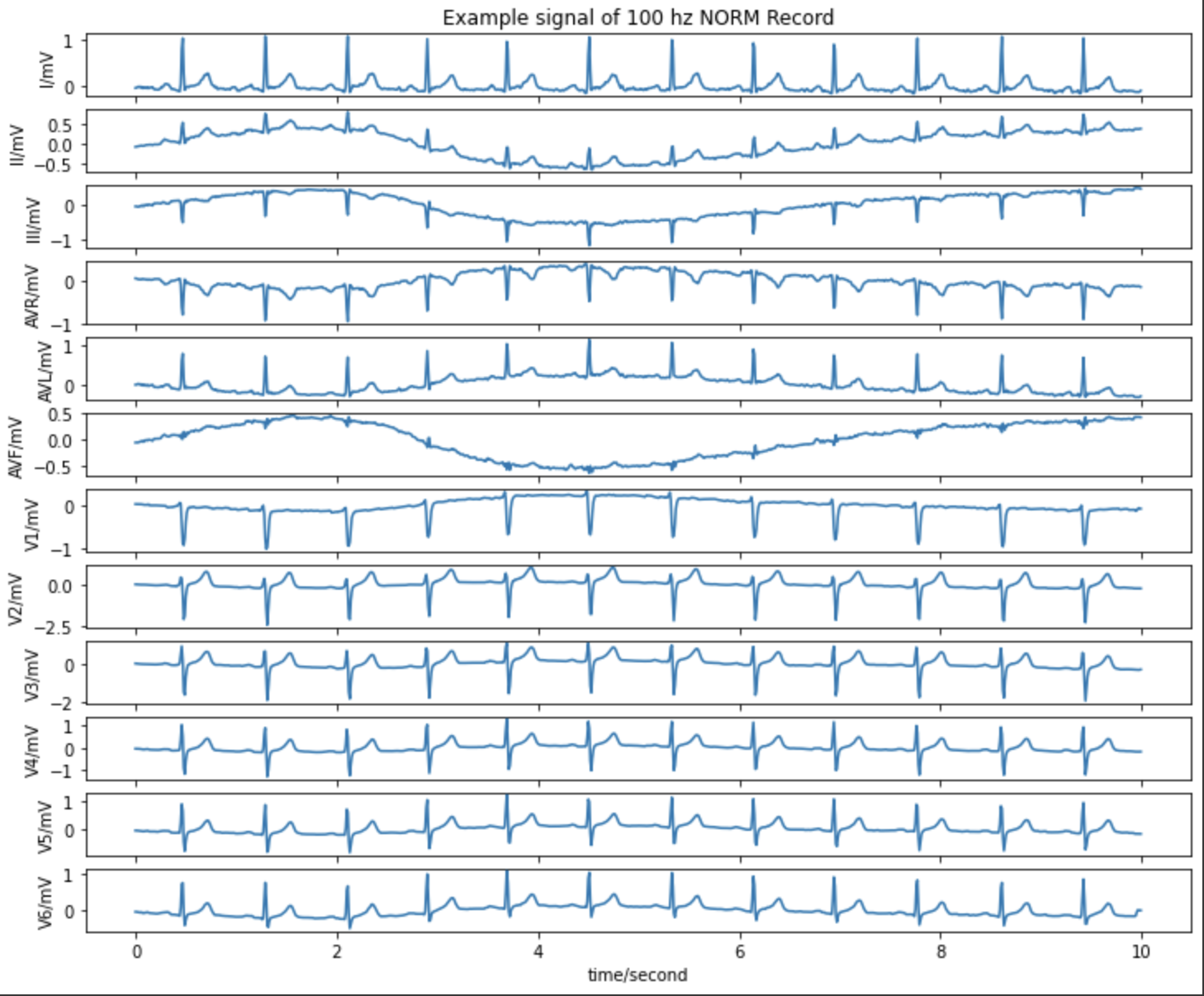}
\end{center}
\caption{12-Lead NORM ECG}
\label{datafig}
\end{figure}

\subsection{Processing clinical notes:}
\begin{figure}[h]
\begin{center}
\includegraphics[width=0.8\textwidth]{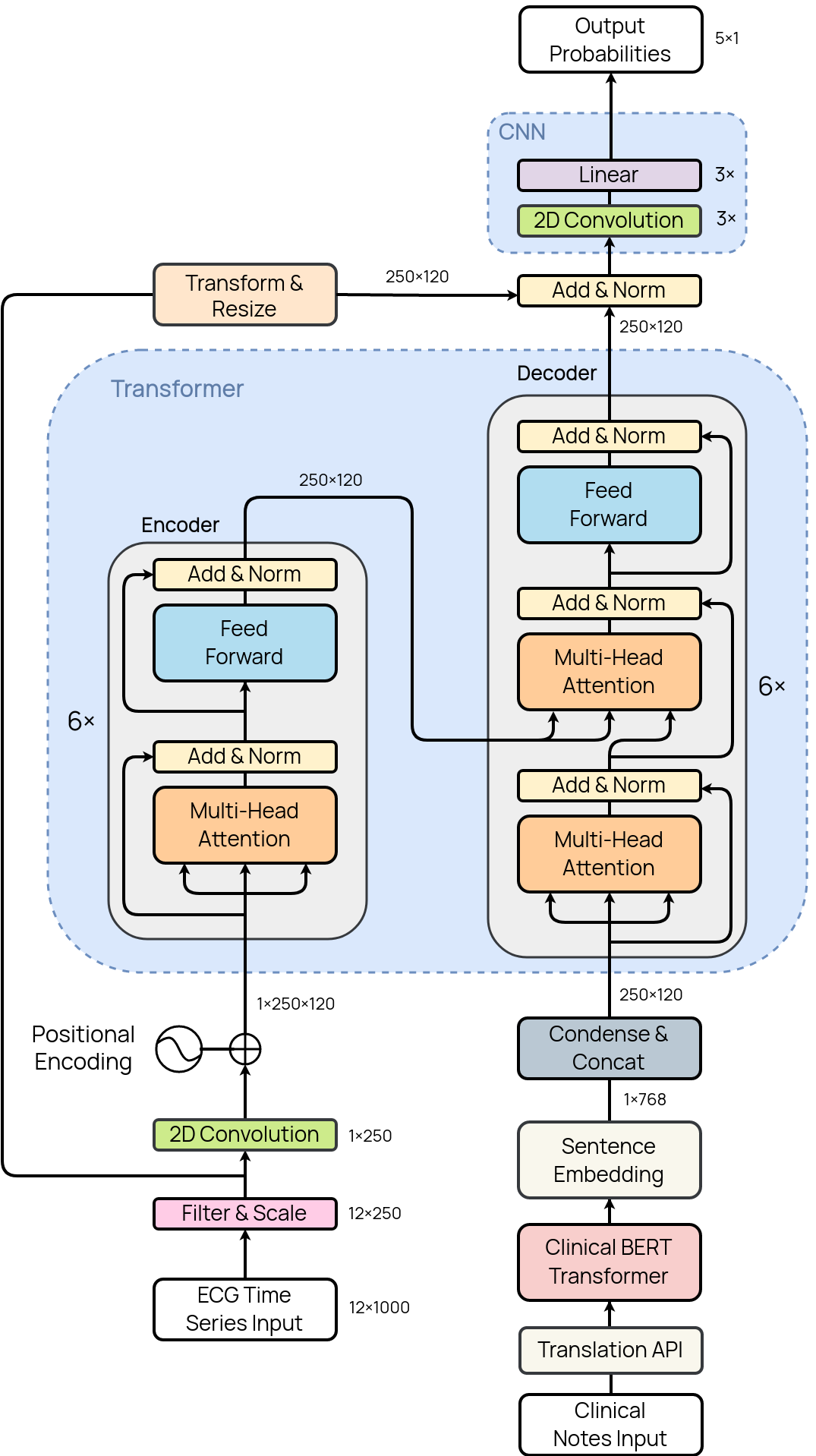}
\end{center}
\caption{MVMTnet Transformer architecture.}
\label{modelfig}
\end{figure}

In order to process this data, the raw data first needs to be loaded and sorted into clinical notes metadata and waveform data. The clinical notes metadata is comprised of information about a patient (age, sex, weight, etc.), as well as two important columns for classification: diagnostic super-classes and report. The diagnostic super-class of a 12-Lead ECG waveform refers to the heart condition(s) which are identified in this patient. The report of each patient is the set of notes written by a doctor documenting the doctor's physical analysis of the patient which is referred to as clinical notes. This metadata can be stored in a pandas data frame so that it is easier to reformat and extract relevant patient information, making sure to include a column for diagnostic super-classes which will contain the labels of the five classes. In examining this column, two problems arise. Firstly, the dataset is skewed, containing roughly $44\%$ of class NORM and multiple labels are assigned to a single ECG. The first issue was solved by under-sampling the data to obtain about $2500$ of each of the five classes effectively cutting the dataset to $12500$ samples. Now that the classes are evenly distributed it made sense to examine the column which contained the clinical notes (reports). This was the main focus for the processing of this metadata and thus all ECGs with a blank report were omitted, cutting the data by a further $1228$ samples. While examining this column, it was observed that $60\%$ of notes were in German. As such, the Google Trans API is used in order to translate the notes to English so that the model can use them to learn. To address the problem of multiple labels, the labels were stored in a list, meaning that a custom dataset class needed to be implemented to allow for multi-hot encoding along with a custom loss function to assess the model’s predictions (see Figure \ref{modelfig}).

\subsection{Processing ECG waveforms:} There are 12 ECG leads recorded for each patient, each with $1000$ data points, sampled at a rate of $100$ Hz; \ref{datafig} shows a normal ECG record. Since the waveforms are highly periodic, the first quarter of the waveform can be used without significant loss of new information. This reduces the number of data points per patient from $12000$ to $3000$, which is computationally favorable. 

An important finding from the examination of ECG graphs is their noisy nature. Significant noise impacts amplitudes and time intervals, which is detrimental to the model's ability to diagnose correctly. Accordingly, a denoising algorithm based on time-frequency filtering was implemented according to Mishra's paper \citep{Mishra2016}. A Daubechies wavelet with 4 vanishing moments was applied to the ECGs, resulting in wavelet coefficients \citep{Addison_2005}. The threshold, calculated as a linear function of the wavelet coefficient median, determined whether to discard or keep specific coefficients. Following soft thresholding on the wavelet coefficients, the inverse wavelet transformation was then applied to reconstruct the denoised ECGs. This process effectively filtered out the high-frequency noise from ECGs while preserving the low-frequency signals \citep{Mishra2016}.

\section{Model Architecture}
\label{Model Architecture}
This novel multi-variate multi-modal Transformer architecture takes inspiration from the core Transformer architecture from the seminal Attention-Is-All-You-Need paper (largely used for NLP tasks), modifying and building around it to incorporate multiple modalities and different embedding types (using transfer learning for clinical notes), and retooling it to handle multi-class classification tasks on time-series signal data \cite{Vaswani2017}.

\subsection{ECG time-series data handler} The network begins with the $12\times250$ block of time-series data passed into the input, which is scaled and filtered for less noise and better interpretability within the model. A 2D convolutional layer is then used to condense the 12 leads into a higher dimension representation (on a per-token basis) of all the leads, to generate a single $1\times250$ sequence to be propagated throughout the rest of the network.

\subsection{Positional encoding} When learning features from sequential (time-series) data, it is critical to be able to maintain information about previous time steps and be aware of a single token’s relative position in the context of the whole sequence. Legacy neural networks such as CNNs treat a sequence of $250$ tokens as $250$ distinct segments, losing any sequential information by doing so. RNNs and LSTMs utilize ‘memory (hidden states)' in the model to keep track of historic tokens and induce sequential feature learning, but they are recursive and slow. Transformers are very effective in modeling time-series data due to their usage of positional encodings, enabling the passage of an entire sequence as a single unit of multiple tokens. The positional encoding layer generates $120\times 1$ embedding vectors for each $10$ms ECG waveform token, facilitating the learning of longer-range trends from how the waveform changes over time, instead of relying on short-range trends from a distinct set of voltage values. As such, the positional encoding layer outputs a $250\times120$ embeddings representation block for the sequence.

\subsection{Self-attention} Self-attention relates different segments in a larger sequence with a degree of "attention" (importance), which helps compute both long and short-range sequential connections. For time-series data, different fluctuations in the sequence may have different "attentions" computed based on which fluctuations have a more out-sized influence on the classification, and these interactions between all the input tokens along with the attention scores are aggregated and passed forward. 

 Multi-headed attention is utilized, which independently runs multiple attention mechanisms in parallel, uses scaled-dot-product attention to pool them, and concatenates and transforms them into the next dimension. Each head takes in three parameters: query, key, and value, which are generally the input passed in three times. Having multiple heads ($12$ heads for the $12$ ECG leads) allows the model to assess different aspects of the sequence differently, just as using $12$ leads of an ECG instead of just a single lead provides much more information about the patient’s overall condition due to the larger ‘surface area’ of signal data \citep{Ribeiro2020}. 

\subsection{Encoder block} Following the generation of positional embeddings, the data is passed into an encoder block consisting of 6 encoder segments. Each of these encoder segments includes a multi-headed attention block, an Add \& Normalize block, followed by a feed-forward (linear) network layer with a final Add \& Normalize block. The encoder stages also use dropout after each layer. These segments are implemented sequentially, with the output of one feeding into the input of the next. The output of the overall encoder is a $250\times120$ block, having attention and positional information embedded in the vectors for each token.

\subsection{Clinical notes modality (Clinical BERT)} Clinical BERT places English language constraints on the input text. Incidentally, German data, in particular German clinical notes, made up $68.7\%$ of the full dataset. Since this represented a significant chunk of data, translation APIs (GoogleTrans API) were utilized to translate all the German clinical notes to English. Although some semantic context was potentially lost through translation, this approach was determined to be more beneficial for the model than simply discarding all the German data.

Transfer learning with Clinical BERT is used to generate sentence embeddings on the clinical notes. All notes were tokenized with Clinical BERT-specific separators, and passed into a pre-trained Clinical BERT model (trained on large datasets of clinical notes for task-specific contextual relevance) \citep{Alsentzer2019}. The output of this Transformer stage is a $1\times768$ sentence embedding vector, which applies to the entire ECG sequence. Since the clinical notes are not time-variant, the vector is condensed and concatenated over the $250$ tokens to generate a $250\times120$ block of those embeddings. This is then passed into the decoder stage along with the outputs of the encoder block. 

\subsection{Decoder} Similar to the encoder block, the decoder block consists of six sequential decoder stages. The clinical notes data modality is processed at the beginning of each of these stages, where the sentence embeddings are taken from the aforementioned clinical notes handler and passed into a multi-head attention and Add \& Normalize layer. Next, for the encoder-decoder attention mechanism, we want to incorporate the attention scores from both the waveform as well as the clinical notes (ensuring the determined influence of each token via attention from both data modalities is considered). So, for the next multi-head attention layer, instead of passing three copies of the ECG encoding outputs into the $q$, $k$, and $v$ parameters, the reshaped sentence embeddings from the clinical notes are passed in as the query parameter. This then goes through the next attention and Add \& Normalize layers. Finally, this passes into a feed-forward (linear) network, and another Add \& Normalize layer. Note that like the encoder, the decoder stages also use dropout after each layer. This sequence is repeated six times with those decoder inputs being passed in at every stage to maintain the positional embeddings throughout the process and ensure that that information is not lost. The output of the decoder is a $250\times120$ sequence\textemdash essentially, the input ECG sequence has been reconstructed and re-represented, but now with embedded attention from both the time-series ECG waveform and clinical notes data modalities, and positional information propagated throughout the Transformer.

\subsection{Output CNN} After the decoder, a residual connection is utilized to add and normalize the original 12-lead input sequences (these are transformed to a $250\times120$ dimension for the Add \& Normalize stage) as well as the output of the Transformer decoder (both in identical formats but representing different qualities of the sequence). This is important since it is necessary to ensure that any ECG signal or attention/positional information that was passed into or generated in the model architecture is not lost and that all the required information is available to pass into the output CNN classification layer. As this is a classification task, the network uses a CNN with $3$ 2D convolutional layers and $3$ linear layers with max-pooling and ReLU. The output of the CNN classifier is a vector of five probabilities (each corresponding to a single cardiac abnormality classification). 

\section{Training}
\label{Training}

\subsection{Optimizer} An Adam optimizer is used, as studies have found that for attention-based techniques, Adam is preferred over stochastic gradient descent (SGD) approaches due to the heavy-tailed distribution of noise present in stochastic gradients from SGD \citep{Zhang2020}. 

\subsection{Loss} From the set of five classes, 30 unique subsets can be formed, which contain all the possible condition states. This is important, as a patient may exhibit more than one condition, and the model should strive to predict multiple conditions correctly. If a patient exhibits multiple conditions, then these conditions are independent of one another. Therefore, a multi-condition prediction has probabilities of each condition that are independent of one another. In order to produce a logical prediction, the output of every neural network has to be put through a sigmoid function, which follows the rule of independent probability, unlike softmax. A loss function that is compatible with multi-label classification is binary cross entropy with logits. The labels are vectors of five classes, where multi-hot encoding represents multiple conditions. Fundamentally, the loss function ‘rewards’ the model when its outputs correctly produce either high probabilities for existing conditions or low probabilities for non-existing conditions.

\subsection{Interpreting accuracy} Accuracy is the cross-reference between the output’s singular condition of highest probability and truth labels. If this condition is matched within the labels, then the model produced an accurate prediction. Otherwise, the prediction was inaccurate. This was the most logical approach to understanding accuracy within the model for several reasons. Although the data is under-sampled to ensure that there is an equivalent class distribution, there were significantly more datasets with singular conditions than with multiple conditions. Therefore, selecting the singular condition of highest probability is sensible for accuracy, which in essence is a binary classifier of an incorrect prediction versus a correct prediction. Additionally, if the model’s strongest prediction is an element of the multi-class labels, then the model is technically accurate. Lastly, the loss function streams the model in the direction of predicting all correspondent classes correctly. In summary, accuracy and loss are representations that differ in one-hot encoding versus multi-hot encoding, respectively.

\section{Results} 
\label{Results} 

\subsection{Quantitative Results}

The best model was observed to perform most accurately under hyperparameter values from Table \ref{hyperparameters}. The model's performance can be quantitatively evaluated using the training curves that compare training loss and validation loss, and training error and validation error.

\begin{figure}[h]
\begin{center}
\includegraphics[width=1.0\textwidth]{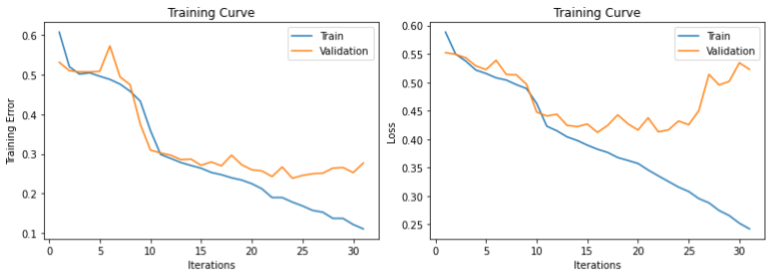}
\end{center}
\caption{MVMTnet optimized training and validation error and loss curves.}
\label{curves}
\end{figure}

From these graphs, the early stopping point before the model starts to overfit is identified as the $24$th epoch with a validation accuracy of $76.16\%$, a notable improvement over all $5$ baselines.

\subsubsection{Optimized MVMTnet Model}
\label{bestmodel}

The tuneable hyperparameter values of the best MVMTnet model are presented in Table \ref{hyperparameters}.

\begin{table}[h]
\caption{Tuneable hyperparameters and correspondent values for the best model.}
\label{hyperparameters}
\begin{tabular}{|c|c|c|c|c|c|c|}
\hline
{\color[HTML]{000000} \textbf{\begin{tabular}[c]{@{}c@{}}Tuneable\\ Hyperparameters\end{tabular}}} & {\color[HTML]{000000} \textbf{\begin{tabular}[c]{@{}c@{}}Learning \\ Rate\end{tabular}}} & {\color[HTML]{000000} \textbf{\begin{tabular}[c]{@{}c@{}}Batch \\ Size\end{tabular}}} & {\color[HTML]{000000} \textbf{\begin{tabular}[c]{@{}c@{}}Model \\ Dimension\end{tabular}}} & {\color[HTML]{000000} \textbf{\begin{tabular}[c]{@{}c@{}}Multi-Head \\ Attention Heads\end{tabular}}} & {\color[HTML]{000000} \textbf{\begin{tabular}[c]{@{}c@{}}Encoder \\ Layers\end{tabular}}} & {\color[HTML]{000000} \textbf{Dropout}} \\ \hline
Best Model & 0.0001 & 4  & 120  & 12  & 6  & 0.2 \\ \hline
\end{tabular}
\end{table}

\subsubsection{Performance on unseen data} The test dataset was obtained during the data procession when the original dataset was being under-sampled. The method which was taken was to find the complement of the under-sampled set, which was an unbalanced set of $11270$ samples. This was then under-sampled further to produce a balanced set of  $1127$ samples containing around $200$ samples of each of the five classes with no empty reports. 

For the MVMTnet model, the obtained test accuracy was 76.03\%, which was within the expected range of the validation accuracy. As such, even on independent test data, our model outperforms all other baseline architectures, rendering it the state-of-the-Art model for this task.

\subsection{Qualitative Results}

\begin{figure}[h]
\begin{center}
\includegraphics[width=1.0\textwidth]{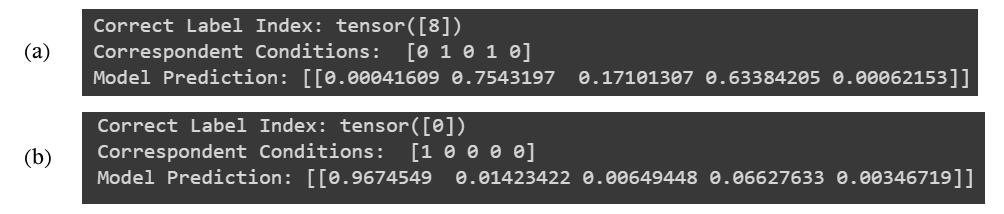}
\end{center}
\caption{Sample outputs of the model; (a) Multi-class label and prediction (b) Singular class label and prediction.}
\label{qual}
\end{figure}

Figure \ref{qual} demonstrates the model's output in regard to two cases within the independent, previously unused test set. The label index is an identifier of the $30$ condition subsets, which directly corresponds to a multi-hot encoding vector of existing conditions. Within (a) of Figure \ref{qual}, the patient exhibits Myocardial Infarction (MI) and Conduction Disturbance (CD). The model's prediction is a vector of class values corresponding to independent probabilities. 

In the case of (a), the model's confidence score is $75.43\%$ for Myocardial Infarction (value at an index of one), and $63.38\%$ for Conduction Disturbance (value at an index of three). More significantly, the confidence scores for non-existing conditions are extremely low with respect to the confidence scores of existing conditions. In the case of (b), the patient exhibits a normal ECG (NORM), as marked by one-hot encoding the condition's label at the index of zero. The confidence score for a normal ECG is $96.75\%$ (value at the index of zero), with negligible confidence scores for other conditions. 

For model outputs that illustrate the model's ability to predict singular conditions as well as multiple conditions with independent confidence scores, refer to \ref{qual}

\subsection{Model Variations and Ablation Analysis }

As MVMTnet is the first multi-variate multi-modal Transformer architecture for multi-class classification, to effectively benchmark its performance and understand the impact of certain core architectural components, we must perform an ablation study and benchmark against variations of MVMTnet (some with components purposely removed, and others with legacy approaches substituted in). Overall, we aim to address the following objectives: (1) the impact of a Transformer architecture in comparison to alternatives like RNN/LSTMs and CNNs, and (2) the benefits of a multi-modality (whether adding an additional data modality such as clinical notes actually improves the performance of the model). As such, we perform an ablation study for MVMTnet to isolate and evaluate the impact of each modular component, using the following uni-modal time-series and multi-modal baseline models as benchmarks: 
(a) CNN (TS-CNN) (b) time-series RNN (TS-RNN) (c) time-series Transformer (TST) (d) multi-modal CNN (M-CNN) (e) multi-modal RNN (M-RNN). 

Among them, TS-CNN, TS-RNN, and TST are stripped down, classical deep learning architectures with the clinical notes modality removed. These can only process a single type of data modality and are intended to help benchmark the addition of the clinical notes modality across multiple neural network architectures, as well as the performance of the Transformer itself over other models in similar conditions, addressing objective 1.

The M-CNN and M-RNN baselines build on the TS-CNN and TS-RNN models by adding the clinical notes BERT component to generate sentence embeddings. They require varying multi-modal data fusion approaches to still perform the same task as MVMTnet, but without the time-series Transformer in the case of M-CNN, and with a time-series RNN swapped in for the Transformer in the case of M-RNN. To do this, M-CNN uses the early concatenation approach, in which the model concatenates the incoming heterogeneous features from different modalities prior to making classifications \citep{Liang2022}. On the other hand, M-RNN takes the early summation approach, where the model takes modality features and weighs them together to merge the data \citep{Liang2022}. This part of the ablation analysis aims to demonstrate the impact of architecture types in a multi-modal setting, addressing objective 1 in the context of objective 2.
\subsubsection{Benchmarking}

To demonstrate the effectiveness of the MVMTnet, we conduct a series of experiments with the baseline models against the optimal model in Section \ref{bestmodel}. The results are given in Table \ref{Accuracies}. Note that since the model is trained on a balanced, high-quality, representative dataset with real-world data, the model's purpose is to make accurate predictions of classifications, and this is a multi-class classification problem, accuracy is a sufficient metric to evaluate the performance of the deep learning models.

\begin{table}[h]
\caption{Performance comparison of MVMTnet with baseline models.}
\label{Accuracies}
\begin{tabular}{|c|c|c|c|c|c|c|}
\hline
{\color[HTML]{000000} \textbf{\begin{tabular}[c]{@{}c@{}} Model\end{tabular}}} & {\color[HTML]{000000} \textbf{TS-CNN}} & {\color[HTML]{000000} \textbf{TS-RNN}} & {\color[HTML]{000000} \textbf{\begin{tabular}[c]{@{}c@{}}TST\end{tabular}}} & {\color[HTML]{000000} \textbf{M-CNN}} & {\color[HTML]{000000} \textbf{M-RNN}} & {\color[HTML]{000000} \textbf{\begin{tabular}[c]{@{}c@{}}MVMTnet\end{tabular}}} \\ \hline
\begin{tabular}[c]{@{}c@{}}Test Accuracies (\%)\end{tabular}& 67.34 &62.77&59.26 &70.39&67.81&76.03\\ \hline
\begin{tabular}[c]{@{}c@{}}Train Accuracies (\%)\end{tabular}&75.60&83.32&88.27 &73.32&74.47&82.17\\ \hline
\begin{tabular}[c]{@{}c@{}}Validation Accuracies (\%)\end{tabular}&69.22&63.87&58.92 &71.11&64.41&76.16\\ \hline
\begin{tabular}[c]{@{}c@{}}Validation Losses (\%)\end{tabular}&0.455&0.607&0.559 &0.430&0.526&0.416\\ \hline
\end{tabular}
\end{table}

The training, validation, and test accuracies from Table \ref{Accuracies} indicate the performance of all uni-modal and multi-modal models. The TS-CNN was the strongest time-series model in comparison to the TS-RNN \& TST models, which appeared to overfit quickly (lack of regularization). Interestingly, the TS-RNN had a higher test accuracy, yet a higher validation loss than the TST; ultimately, the TS-RNN made more correct predictions, but with weaker confidence scores. Conversely, the multi-modal models exhibited improvements in accuracies from their uni-modal counterparts; this was rather expected. Referencing the test accuracies: the M-CNN \& M-RNN experienced 3-5\% boosts from their time-series counterparts, while the MVMTnet model saw a 16.77\% increase from the TST. The explanation lies in modality fusion. The M-RNN \& M-CNN models fused modalities through early concatenation, while the MVMTnet used cross-attention from a Transformer layer.
\section{Discussion}

\subsection{Sensitivity of ECG signal data} The interference of noisy data must be considered as a potential cause of misclassification errors. The variability induced by such artifacts makes ECG signal classification an inherently difficult task and requires denoising via a Daubechies wavelet during data pre-processing as shown in Section \ref{Data Processing}. Despite undertaking this step, there exists a possibility that there is still a non-negligible amount of undesirable signals left in the ECG signal data, which can create an obstacle to making a true diagnostic prediction. This further supports the case that the additional modality of the clinical notes is needed to corroborate such predictions.

\subsection{Learning observations}

\subsubsection{Error and loss inverse correlation} As defined in Section \ref{Training}, calculating loss consisted of performing cross entropy with logits against multi-hot encoded labels. Oppositely, calculating accuracy consisted of selecting a singular condition of highest probability and cross-referencing it to the labels. Within the model, certain epoch intervals exhibited an increasing validation loss with a decrease in training error. In most models, this is observed as overfitting. However, due to the nature of this model's aforementioned definitions of loss and accuracy, a more relevant measure of overfitting is the trend in validation error, cross-referenced with the loss trends, where as long as the validation error continues to decrease, there is no need to define an early stopping point to prevent overfitting.

\subsubsection{Impact of learning rate on updating weights} The learning rate impacts the scale of the model's response to error, which is directly proportional to how drastically the weights are changed. A very small learning rate ($0.0001$) was used in this model because it was observed that the model was beginning to memorize a large dataset very quickly. This learning rate allowed for validation loss and error to decrease at smaller steps, ultimately resulting in the model neither over-fitting nor under-fitting within the desired range of $20$ epochs. However, lowering the learning rate slowed training, which negatively impacted the resource limitations and GPU constraints.

\subsection{Multi-variate multi-headed attention} 

\begin{figure}[h]
  \centering
  \subfloat[Multi-variate multi-headed attention layer.]{\includegraphics[width=0.45\textwidth]{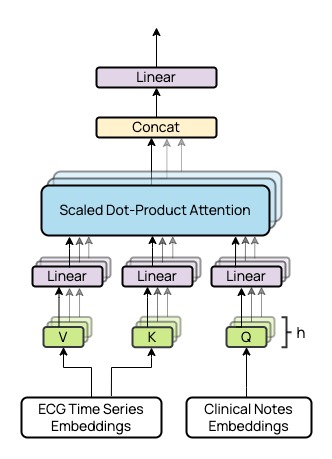}}
  \hfill
  \subfloat[Normal multi-headed attention layer. \citep{Vaswani2017}]{\includegraphics[width=0.45\textwidth]{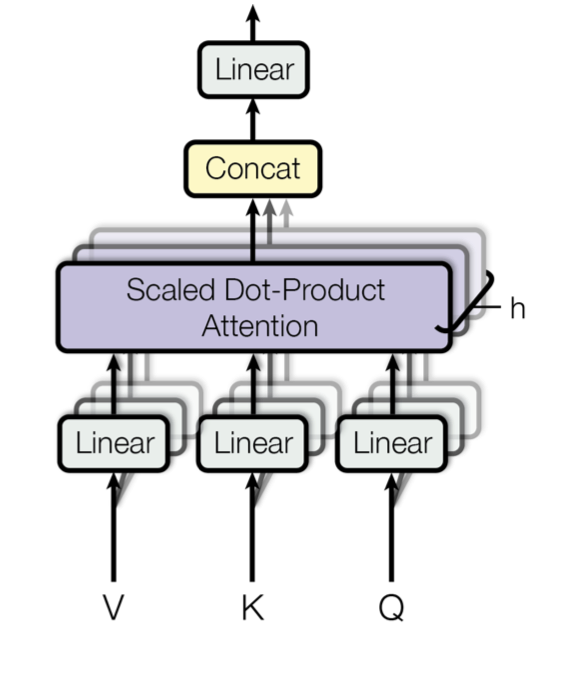}}
  \caption{Difference between multi-headed attention layers.}
  \label{multiheadlayer}
\end{figure}

During the design process, one of the goals was to explore the potential for developing a model that could not only provide an effective analysis of the provided data but also provide clinically-relevant insights based on the outputs of different parts of the model. In this case, we wanted to see if clinically relevant insights could be drawn from how the outputs of the $12$ heads were pooled and condensed into a single representation. Although our final architecture uses the traditional multi-headed attention layer which linearly projects a single query, key, and value triplet $h$ (number of heads) times with distinct linear projections that were learned during training \citep{Vaswani2017}, we wanted to architecturally enforce the assignment of each ECG lead to each head in this layer. That way, as seen in Figure \ref{multiheadlayer}, instead of passing in a single $q$, $k$, and $v$ triplet and linearly projecting from there, we would isolate the waveform for each lead from the beginning, run $12$ independent encoder blocks for the $12$ leads, and then run a multi-variate multi-head attention layer which would directly take the embeddings generated from the $12$ independent encoders instead of linearly projecting them from a single, already-condensed set of embeddings. This would then allow us to compare the output vectors of each head and use techniques like cosine similarity indexes (or any similar/alternative vector comparison approach) and potentially examine how much the features learned from certain heads/leads were carried forward and maintained in the pooled representation (in this case, we might start with the assumption that the more similar a head vector is to the pooled vector, the more influential it was or the more information from that was maintained, although this assumption would change based on the pooling approach used). While we did implement this approach, we were unable to effectively train our architecture using this approach due to the massive computational load of running 12 independent encoders for each lead waveform prior to feeding it into the multi-variate multi-head attention layer in the decoder (12x the computation and training time as what is currently required for all model layers up to the decoder). As such, we leave this for experimentation in future iterations of research with this or similar architectures.

\subsection{Further analysis of multi-headed attention for 12-lead ECGs}
\begin{figure}[h]
  \centering
  {\includegraphics[width=1\textwidth]{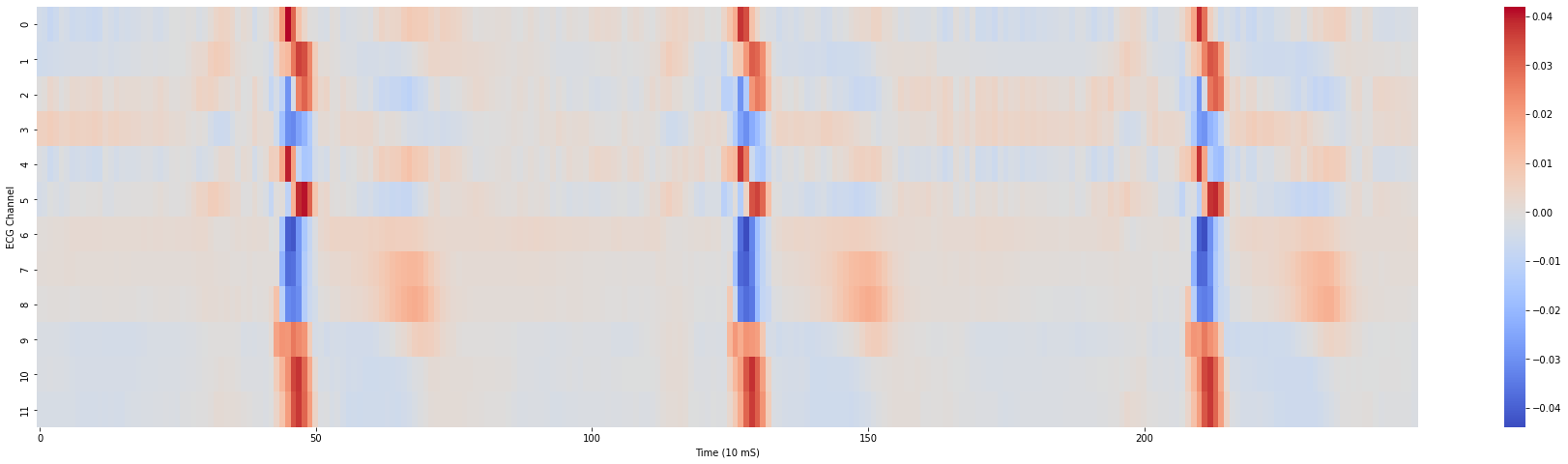}}
  \caption{Color-coded visual representation of signal data corresponding to 12 ECG Leads.}
  \label{heatmap}
\end{figure}

There were certain experiments considered to explore the possibility of deriving clinically-relevant insights from the model's feature learning, with some being more robust than others. This included an attention-based visualization via a heat map that demonstrated what segments of the sequence the model attends to, as observed in Figure \ref{heatmap}. The heat map was obtained by pooling the attention head weights and performing matrix multiplication with the input 12-lead ECG within the encoder layer. Evidently, the model attends in 80-120 time segments. This periodic attending predictably occurs because of voltage peaks and troughs, which are distinctive within the waveform. Additionally, it appears to be that the model is indifferent (pays less attention) to voltage segments that are non-changing. This attention dynamic seems to be consistent with what one may or may not pay attention to from a waveform. Unfortunately, these color-coded signal maps often do not lead to significant conclusions, and as it is in this case, not much can be concluded, other than that the model attends more dramatically to firing segments of the ECG. Although further analysis can be done on the way the Transformer calculates attention for different parts of the waveform for different leads and how that varies based on the classification, the analysis done as a part of this work does not indicate any new clinically-relevant insights from attention other than what is already known about ECG waveforms.

\section{Ethical Considerations}
Ethical considerations are critical especially when sensitive, personal information such as medical patient notes are in question. Care must be taken to ensure patient data is rinsed of any personally identifying tags, and that the data used cannot be traced back to an individual for privacy purposes. This is especially important with the advent of wearable technologies that make physiological data capture and analysis far more accessible—using monitoring/predictive models like ours on personal devices, meaning that such applications must be careful about how they handle metadata for patient data being processed. 

Additionally, with regard to the deployment of models such as ours in critical settings like hospitals, AI systems aren't foolproof and have a not-so-insignificant margin of error. There is always a certain risk of misdiagnosis, so there are discussions to be had about how much weight should be placed on results from an AI model in the field, who/what is liable for circumstances surrounding AI-based decisions, and how these models fit in long term in an often-changing medical environment.

\section{Conclusion}
In this work, we presented MVMTnet, a novel multi-variate multi-modal Transformer architecture that can perform multi-class classification of cardiac abnormalities using 12-lead ECG waveform data as well as clinical notes. We have demonstrated its superior performance to other traditional deep learning approaches, reaffirming the choice of Transformer architecture and multi-modal implementation for this particular use case, and establishing it as the state-of-the-art for this task. 

And, although Transformers have historically been used for natural-language-processing tasks \citep{Lin2022}, our successful implementation on medical time-series data showcases the efficacy of Transformers across multiple application domains. Additionally, our successful use of transfer learning to integrate the clinical notes modality and improve the accuracy of all tested models further showcases the benefits of transfer learning in cross-domain applications, especially in multi-modal settings where both transfer learning and supervised learning are utilized with different Transformers within the model.

The potential applications of models such as ours are wide-ranging, ranging from deploying such models out in the field (real-time monitoring, faster response time, greater assistance for on-site medical staff), in our daily lives (using data from wearable technologies like smartwatches for health monitoring and risk-prediction), as well as in controlled research settings (organoid research; signal modeling, forecasting, and classification; and even drug efficacy testing). We are excited by the potential of MVMTnet and plan to not only further develop the model architecture, but also explore the application space and work towards bridging the gap between the research and deployment of such clinical AI models in real life.

Our codebase can be found at: \href{https://www.github.com/Ankur-Samanta/MVMTnet-Transformer_ECGClassification}{MVMTnet Codebase}.

\section{Disclosure of Potential Conflicts of Interest}
The authors declare that there are no conflicts of interest regarding this work.

\label{last_page}

\newpage
\bibliography{mvmtnet}

\begin{thebibliography}{27}
\providecommand{\natexlab}[1]{#1}
\providecommand{\url}[1]{\texttt{#1}}
\expandafter\ifx\csname urlstyle\endcsname\relax
  \providecommand{\doi}[1]{doi: #1}\else
  \providecommand{\doi}{doi: \begingroup \urlstyle{rm}\Url}\fi

\bibitem[Adams \& Choi(2012)Adams and Choi]{Adams2012}
E.~Roland Adams and Anthony Choi.
\newblock Using neural networks to predict cardiac arrhythmias.
\newblock pp.\  402--407, 2012.
\newblock \doi{10.1109/ICSMC.2012.6377734}.

\bibitem[Addison(2005)]{Addison_2005}
Paul~S Addison.
\newblock Wavelet transforms and the ecg: a review.
\newblock \emph{Physiological Measurement}, 26\penalty0 (5):\penalty0 R155, aug
  2005.
\newblock \doi{10.1088/0967-3334/26/5/R01}.
\newblock URL \url{https://dx.doi.org/10.1088/0967-3334/26/5/R01}.

\bibitem[Ahmad et~al.(2021)Ahmad, Tabassum, Guan, and Khan]{ahmad2021}
Zeeshan Ahmad, Anika Tabassum, Ling Guan, and Naimul~Mefraz Khan.
\newblock Ecg heartbeat classification using multimodal fusion.
\newblock \emph{IEEE Access}, 9:\penalty0 100615--100626, 2021.

\bibitem[Alsentzer et~al.(2019)Alsentzer, Murphy, Boag, Weng, Jin, Naumann, and
  McDermott]{Alsentzer2019}
Emily Alsentzer, John~R. Murphy, Willie Boag, Wei-Hung Weng, Di~Jin, Tristan
  Naumann, and Matthew B.~A. McDermott.
\newblock Publicly available clinical bert embeddings.
\newblock 2019.
\newblock \doi{10.48550/ARXIV.1904.03323}.
\newblock URL \url{https://arxiv.org/abs/1904.03323}.

\bibitem[Bizopoulos \& Koutsouris(2018)Bizopoulos and
  Koutsouris]{bizopoulos2018}
Paschalis Bizopoulos and Dimitrios Koutsouris.
\newblock Deep learning in cardiology.
\newblock \emph{IEEE reviews in biomedical engineering}, 12:\penalty0 168--193,
  2018.

\bibitem[Chamatidis et~al.(2017)Chamatidis, Katsika, and
  Spathoulas]{chamatidis2017}
Ilias Chamatidis, Aggeliki Katsika, and Georgios Spathoulas.
\newblock Using deep learning neural networks for ecg based authentication.
\newblock In \emph{2017 international Carnahan conference on security
  technology (ICCST)}, pp.\  1--6. IEEE, 2017.

\bibitem[Deznabi et~al.(2021)Deznabi, Iyyer, and Fiterau]{deznabi2021}
Iman Deznabi, Mohit Iyyer, and Madalina Fiterau.
\newblock Predicting in-hospital mortality by combining clinical notes with
  time-series data.
\newblock In \emph{Findings of the Association for Computational Linguistics:
  ACL-IJCNLP 2021}, pp.\  4026--4031, 2021.

\bibitem[Ebrahimi et~al.(2020)Ebrahimi, Loni, Daneshtalab, and
  Gharehbaghi]{ebrahimi2020}
Zahra Ebrahimi, Mohammad Loni, Masoud Daneshtalab, and Arash Gharehbaghi.
\newblock A review on deep learning methods for ecg arrhythmia classification.
\newblock \emph{Expert Systems with Applications: X}, 7:\penalty0 100033, 2020.
\newblock ISSN 2590-1885.
\newblock \doi{https://doi.org/10.1016/j.eswax.2020.100033}.
\newblock URL
  \url{https://www.sciencedirect.com/science/article/pii/S2590188520300123}.

\bibitem[Eltrass et~al.(2022)Eltrass, Tayel, and Ammar]{Eltrass2022}
Ahmed~S. Eltrass, Mazhar~B. Tayel, and Abeer~I. Ammar.
\newblock {Automated ECG multi-class classification system based on combining
  deep learning features with HRV and ECG measures}.
\newblock \emph{Neural Computing and Applications}, 34\penalty0 (11):\penalty0
  8755--8775, 2022.
\newblock ISSN 1433-3058.
\newblock \doi{10.1007/s00521-022-06889-z}.
\newblock URL \url{https://doi.org/10.1007/s00521-022-06889-z}.

\bibitem[Hatami et~al.(2022)Hatami, Cho, Mechtouff, Eker, Rousseau, and
  Frindel]{Hatami2022}
Nima Hatami, Tae-Hee Cho, Laura Mechtouff, Omer~Faruk Eker, David Rousseau, and
  Carole Frindel.
\newblock Cnn-lstm based multimodal mri and clinical data fusion for predicting
  functional outcome in stroke patients.
\newblock 2022.
\newblock \doi{10.48550/ARXIV.2205.05545}.
\newblock URL \url{https://arxiv.org/abs/2205.05545}.

\bibitem[Isin \& Ozdalili(2017)Isin and Ozdalili]{isin2017}
Ali Isin and Selen Ozdalili.
\newblock Cardiac arrhythmia detection using deep learning.
\newblock \emph{Procedia computer science}, 120:\penalty0 268--275, 2017.

\bibitem[Jo et~al.(2019)Jo, Nho, and Saykin]{jo2019}
Taeho Jo, Kwangsik Nho, and Andrew~J Saykin.
\newblock Deep learning in alzheimer's disease: diagnostic classification and
  prognostic prediction using neuroimaging data.
\newblock \emph{Frontiers in aging neuroscience}, 11:\penalty0 220, 2019.

\bibitem[Khadanga et~al.(2019)Khadanga, Aggarwal, Joty, and
  Srivastava]{Khadanga2019}
Swaraj Khadanga, Karan Aggarwal, Shafiq~R. Joty, and Jaideep Srivastava.
\newblock Using clinical notes with time series data for {ICU} management.
\newblock \emph{CoRR}, abs/1909.09702, 2019.
\newblock URL \url{http://arxiv.org/abs/1909.09702}.

\bibitem[Liang et~al.(2022)Liang, Tohti, and Hamdulla]{Liang2022}
Yi~Liang, Turdi Tohti, and Askar Hamdulla.
\newblock Multimodal false information detection method based on text-cnn and
  se module.
\newblock \emph{PloS one}, 17:\penalty0 e0277463, 11 2022.
\newblock \doi{10.1371/journal.pone.0277463}.

\bibitem[Lin et~al.(2022)Lin, Wang, Liu, and Qiu]{Lin2022}
Tianyang Lin, Yuxin Wang, Xiangyang Liu, and Xipeng Qiu.
\newblock A survey of transformers.
\newblock \emph{AI Open}, 3:\penalty0 111--132, 2022.
\newblock ISSN 2666-6510.
\newblock \doi{https://doi.org/10.1016/j.aiopen.2022.10.001}.
\newblock URL
  \url{https://www.sciencedirect.com/science/article/pii/S2666651022000146}.

\bibitem[Lui \& Chow(2018)Lui and Chow]{Lui2018}
Hin~Wai Lui and King~Lau Chow.
\newblock Multiclass classification of myocardial infarction with convolutional
  and recurrent neural networks for portable ecg devices.
\newblock \emph{Informatics in Medicine Unlocked}, 13:\penalty0 26--33, 2018.
\newblock ISSN 2352-9148.
\newblock \doi{https://doi.org/10.1016/j.imu.2018.08.002}.
\newblock URL
  \url{https://www.sciencedirect.com/science/article/pii/S2352914818301333}.

\bibitem[Mishra et~al.(2016)Mishra, Singh, and Sahu]{Mishra2016}
Ankita Mishra, Ashutosh~Kumar Singh, and Sitanshu~Sekhar Sahu.
\newblock Ecg signal denoising using time-frequency based filtering approach.
\newblock \emph{2016 International Conference on Communication and Signal
  Processing (ICCSP)}, pp.\  0503--0507, 2016.
\newblock \doi{10.1109/ICCSP.2016.7754188}.

\bibitem[Park et~al.(2022)Park, An, Kim, Jung, Gil, Jang, Lee, and young
  Oh]{park2022}
Junsang Park, Junho An, Jinkook Kim, Sunghoon Jung, Yeongjoon Gil, Yoojin Jang,
  Kwanglo Lee, and Il~young Oh.
\newblock Study on the use of standard 12-lead ecg data for rhythm-type ecg
  classification problems.
\newblock \emph{Computer Methods and Programs in Biomedicine}, 214:\penalty0
  106521, 2022.
\newblock ISSN 0169-2607.
\newblock \doi{https://doi.org/10.1016/j.cmpb.2021.106521}.
\newblock URL
  \url{https://www.sciencedirect.com/science/article/pii/S0169260721005952}.

\bibitem[Ribeiro et~al.(2020)Ribeiro, Ribeiro, Paixão, Oliveira, Gomes,
  Canazart, Ferreira, Andersson, Macfarlane, Jr, Schön, and
  Ribeiro]{Ribeiro2020}
Antonio~H. Ribeiro, Manoel~Horta Ribeiro, Gabriela Paixão, Derick Oliveira,
  Paulo Gomes, Jéssica Canazart, Milton Ferreira, Carl Andersson, Peter
  Macfarlane, Wagner~Meira Jr, Thomas Schön, and Antonio~Luiz Ribeiro.
\newblock Automatic diagnosis of the 12-lead ecg using a deep neural network.
\newblock \emph{Nature Communications}, 11, 04 2020.
\newblock \doi{10.1038/s41467-020-15432-4}.

\bibitem[S \& K(2022)S and K]{Chandru2022}
Chandru~A. S and Seetharam K.
\newblock Processing of clinical notes for efficient diagnosis with dual lstm.
\newblock \emph{International Journal of Advanced Computer Science and
  Applications}, 13\penalty0 (2), 2022.
\newblock \doi{10.14569/IJACSA.2022.0130247}.
\newblock URL \url{http://dx.doi.org/10.14569/IJACSA.2022.0130247}.

\bibitem[Sherstinsky(2018)]{Sherstinsky2018}
Alex Sherstinsky.
\newblock Fundamentals of recurrent neural network {(RNN)} and long short-term
  memory {(LSTM)} network.
\newblock \emph{CoRR}, abs/1808.03314, 2018.
\newblock URL \url{http://arxiv.org/abs/1808.03314}.

\bibitem[Smith et~al.(2019)Smith, Rapin, Li, Fleureau, Fennell, Walsh, Rosier,
  Fiorina, and Gardella]{Smith2019}
Stephen~W. Smith, Jeremy Rapin, Jia Li, Yann Fleureau, William Fennell,
  Brooks~M. Walsh, Arnaud Rosier, Laurent Fiorina, and Christophe Gardella.
\newblock A deep neural network for 12-lead electrocardiogram interpretation
  outperforms a conventional algorithm, and its physician overread, in the
  diagnosis of atrial fibrillation.
\newblock \emph{IJC Heart \& Vasculature}, 25:\penalty0 100423, 2019.
\newblock ISSN 2352-9067.
\newblock \doi{https://doi.org/10.1016/j.ijcha.2019.100423}.
\newblock URL
  \url{https://www.sciencedirect.com/science/article/pii/S2352906719301241}.

\bibitem[Tan et~al.(2017)Tan, Sun, Zhang, Chen, and Liu]{tan2017}
Chuanqi Tan, Fuchun Sun, Wenchang Zhang, Jianhua Chen, and Chunfang Liu.
\newblock Multimodal classification with deep convolutional-recurrent neural
  networks for electroencephalography.
\newblock In \emph{Neural Information Processing: 24th International
  Conference, ICONIP 2017, Guangzhou, China, November 14-18, 2017, Proceedings,
  Part II 24}, pp.\  767--776. Springer, 2017.

\bibitem[Vaswani et~al.(2017)Vaswani, Shazeer, Parmar, Uszkoreit, Jones, Gomez,
  Kaiser, and Polosukhin]{Vaswani2017}
Ashish Vaswani, Noam Shazeer, Niki Parmar, Jakob Uszkoreit, Llion Jones,
  Aidan~N. Gomez, Lukasz Kaiser, and Illia Polosukhin.
\newblock Attention is all you need.
\newblock \emph{CoRR}, abs/1706.03762, 2017.
\newblock URL \url{http://arxiv.org/abs/1706.03762}.

\bibitem[Wagner et~al.(2020)Wagner, Strodthoff, Bousseljot, Samek, and
  Schaeffter]{Wagner2020}
Patrick Wagner, Nils Strodthoff, Ralf-Dieter Bousseljot, Wojciech Samek, and
  Tobias Schaeffter.
\newblock Ptb-xl, a large publicly available electrocardiography dataset
  (version 1.0.1).
\newblock 2020.
\newblock URL \url{https://doi.org/10.13026/x4td-x982}.

\bibitem[Zerveas et~al.(2020)Zerveas, Jayaraman, Patel, Bhamidipaty, and
  Eickhoff]{Zerveas2020}
George Zerveas, Srideepika Jayaraman, Dhaval Patel, Anuradha Bhamidipaty, and
  Carsten Eickhoff.
\newblock A transformer-based framework for multivariate time series
  representation learning.
\newblock 2020.
\newblock \doi{10.48550/ARXIV.2010.02803}.
\newblock URL \url{https://arxiv.org/abs/2010.02803}.

\bibitem[Zhang et~al.(2020)Zhang, Yin, Zeng, Yuan, and Zhang]{Zhang2020}
D~Zhang, C~Yin, J~Zeng, X~Yuan, and P~Zhang.
\newblock Combining structured and unstructured data for predictive models: a
  deep learning approach.
\newblock \emph{BMC medical informatics and decision making}, 20, 2020.
\newblock \doi{10.1186/s12911-020-01297-6}.

\end{thebibliography}
\bibliographystyle{iclr2022_conference}

\newpage
\section*{Appendix}
\begin{figure}[h]
\begin{center}
\includegraphics[width=0.8\textwidth]{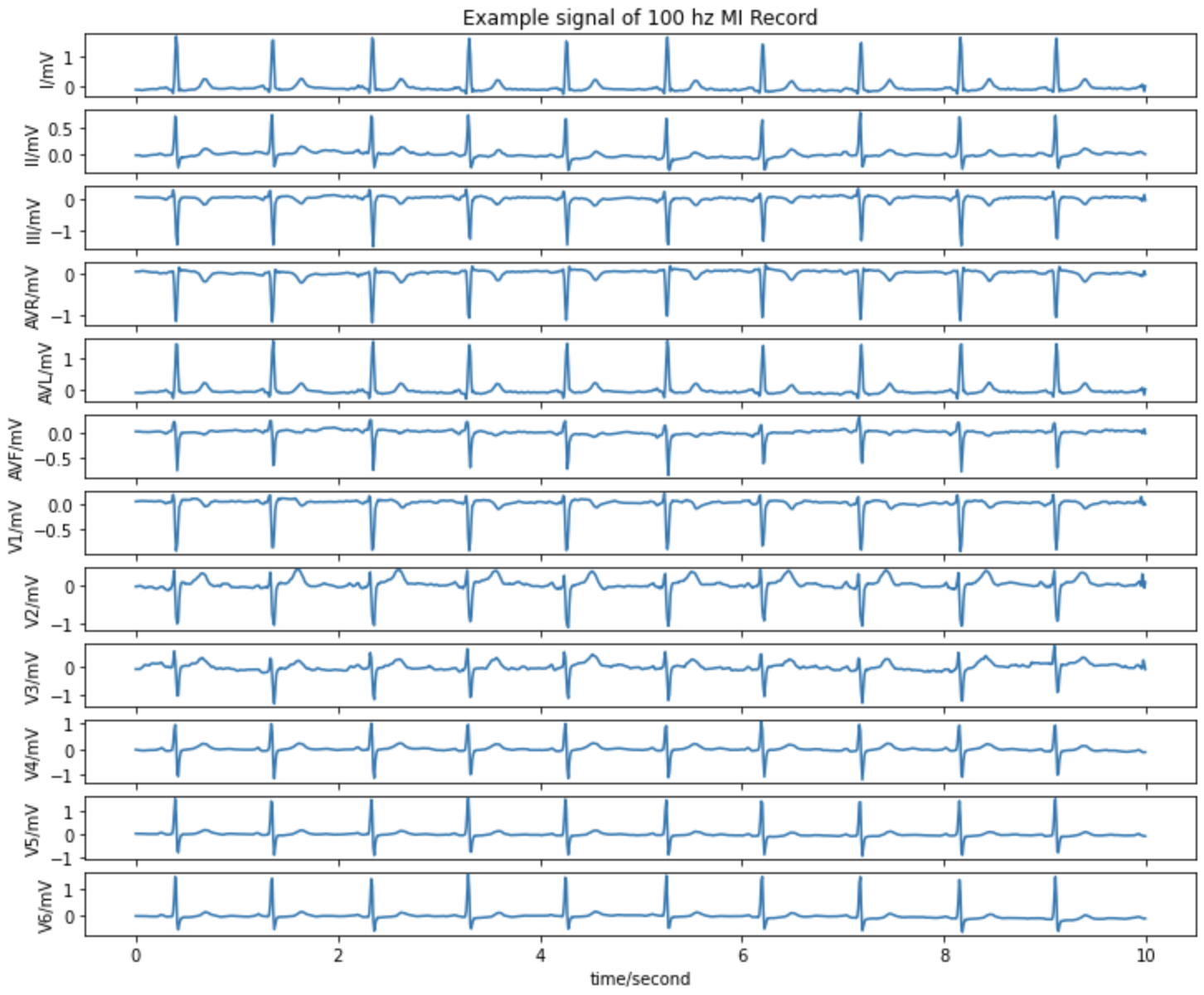}
\end{center}
\caption{12-Lead MI ECG}
\label{test}
\end{figure}

\begin{figure}[h]
\begin{center}
\includegraphics[width=0.8\textwidth]{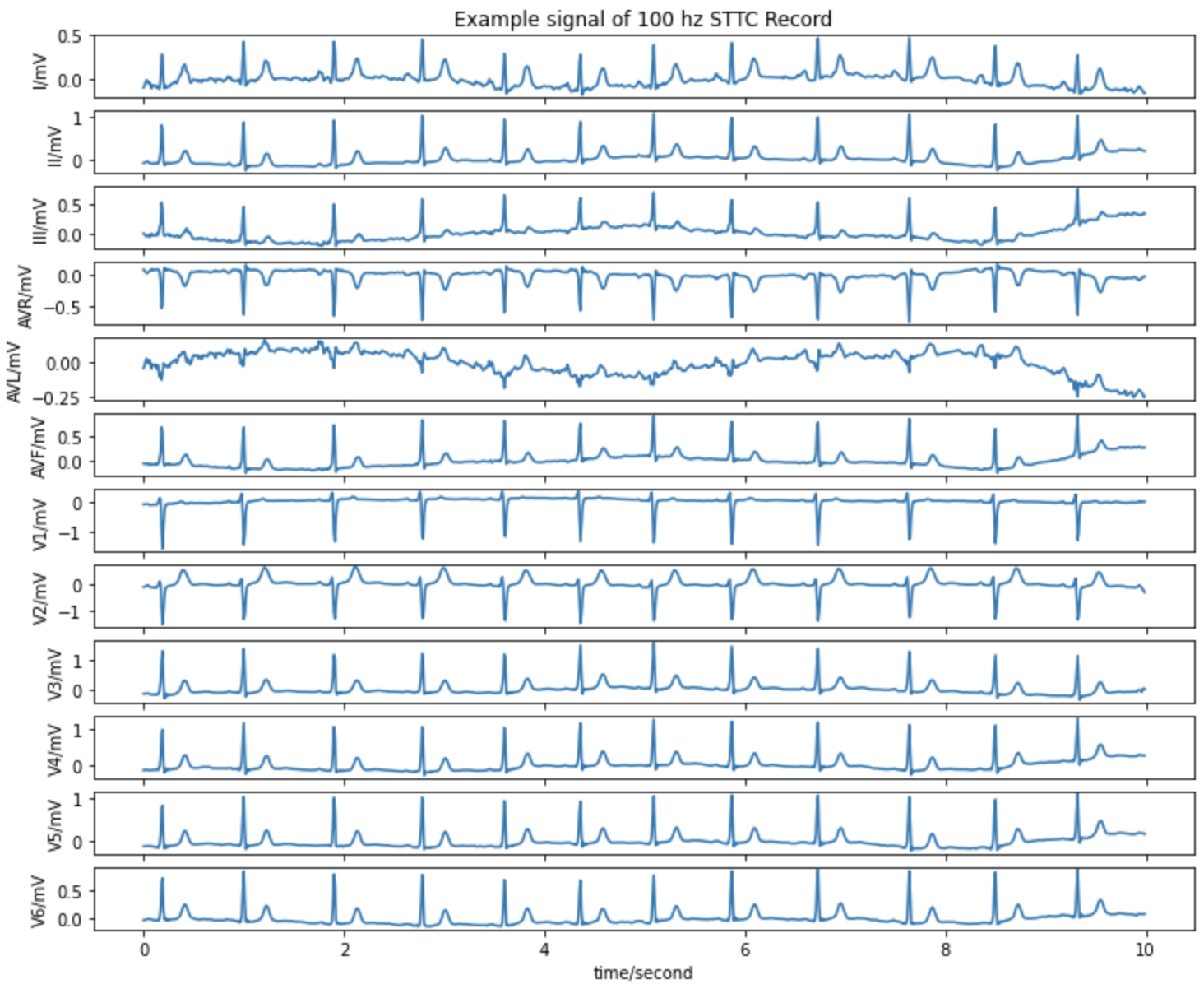}
\end{center}
\caption{12-Lead STTC ECG}
\label{test}
\end{figure}

\begin{figure}[h]
\begin{center}
\includegraphics[width=0.8\textwidth]{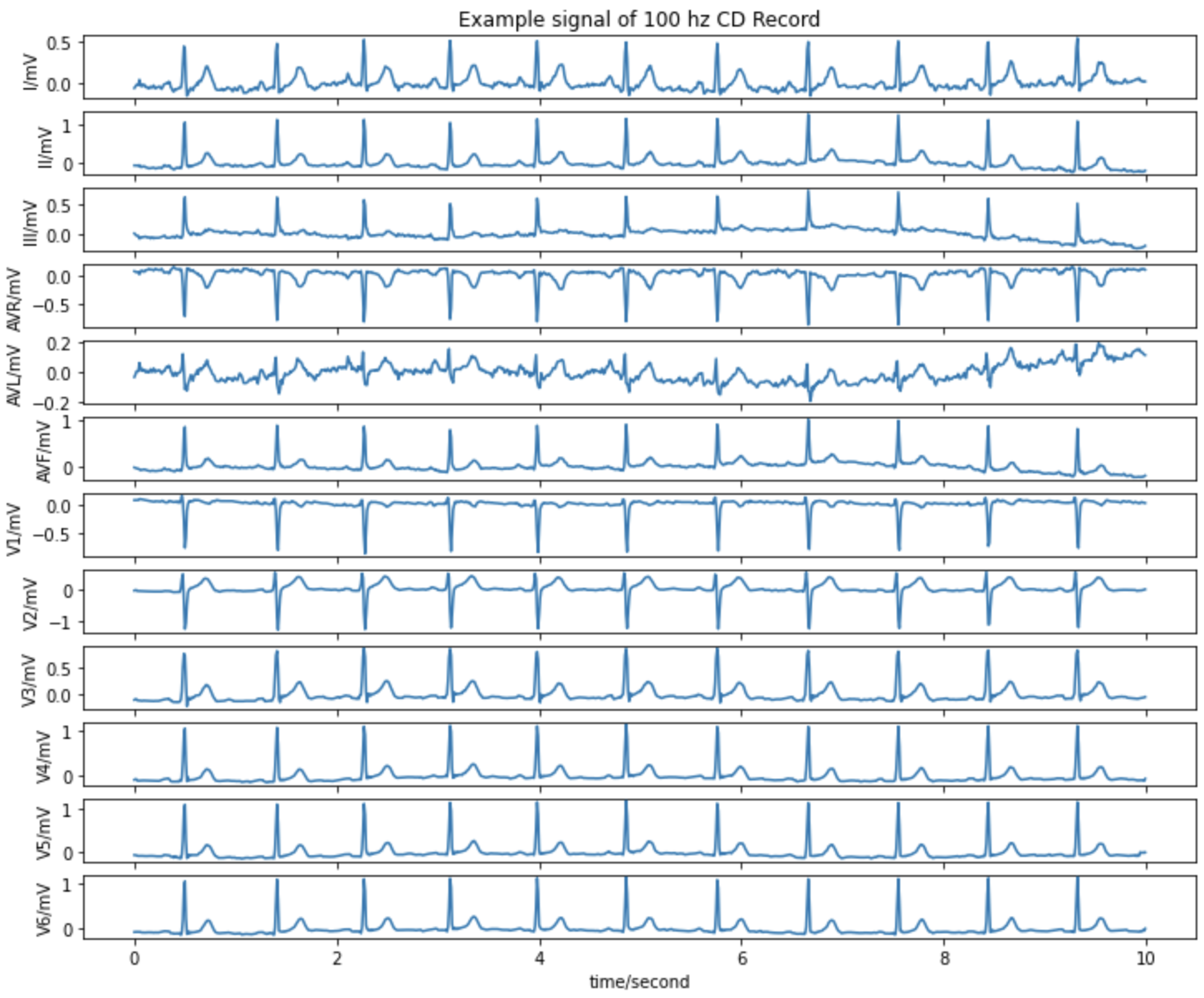}
\end{center}
\caption{12-Lead CD ECG}
\label{test}
\end{figure}

\begin{figure}[h]
\begin{center}
\includegraphics[width=0.8\textwidth]{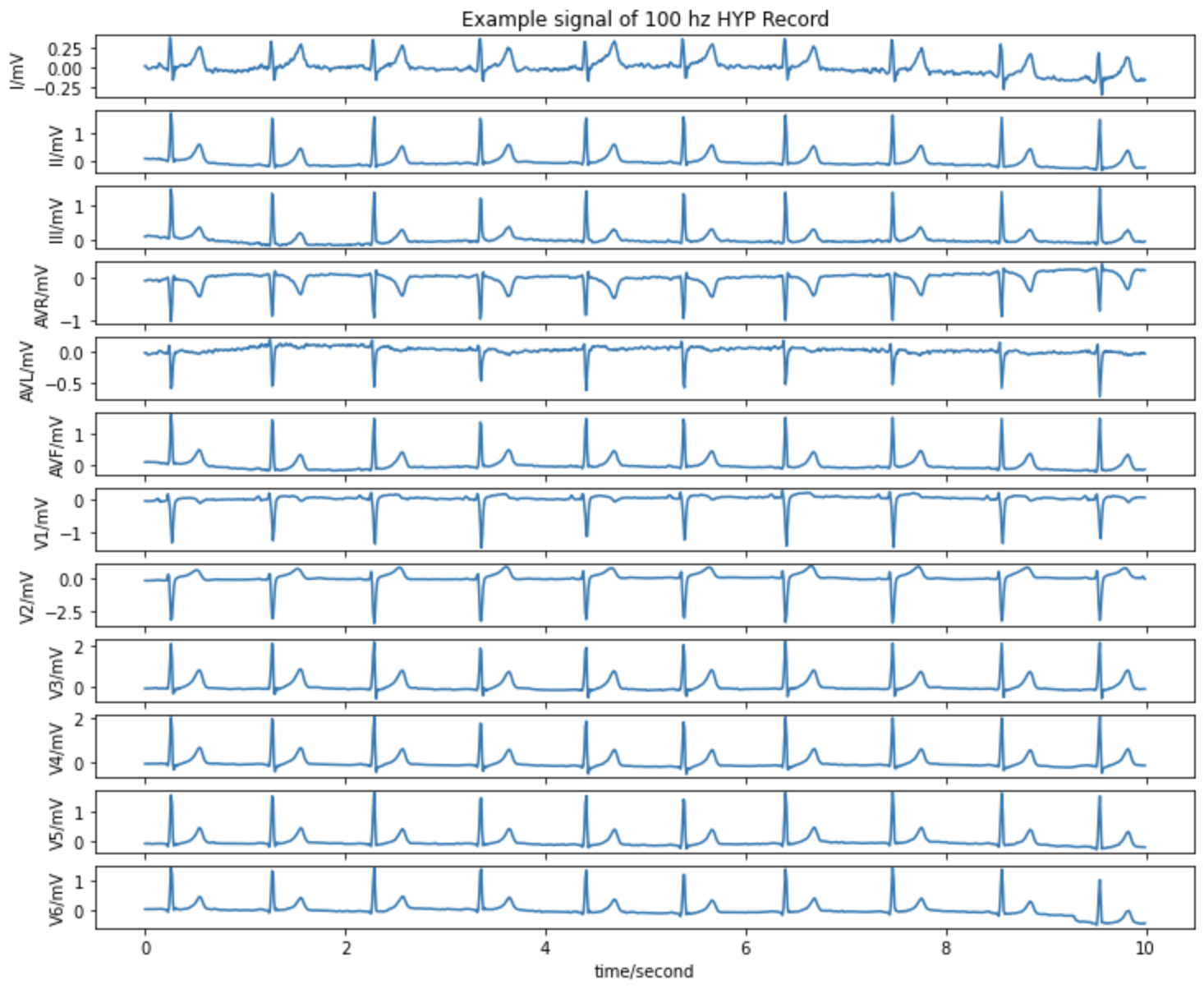}
\end{center}
\caption{12-Lead HYP ECG}
\label{test}
\end{figure}

\begin{figure}[h]
\begin{center}
\includegraphics[width=0.9\textwidth]{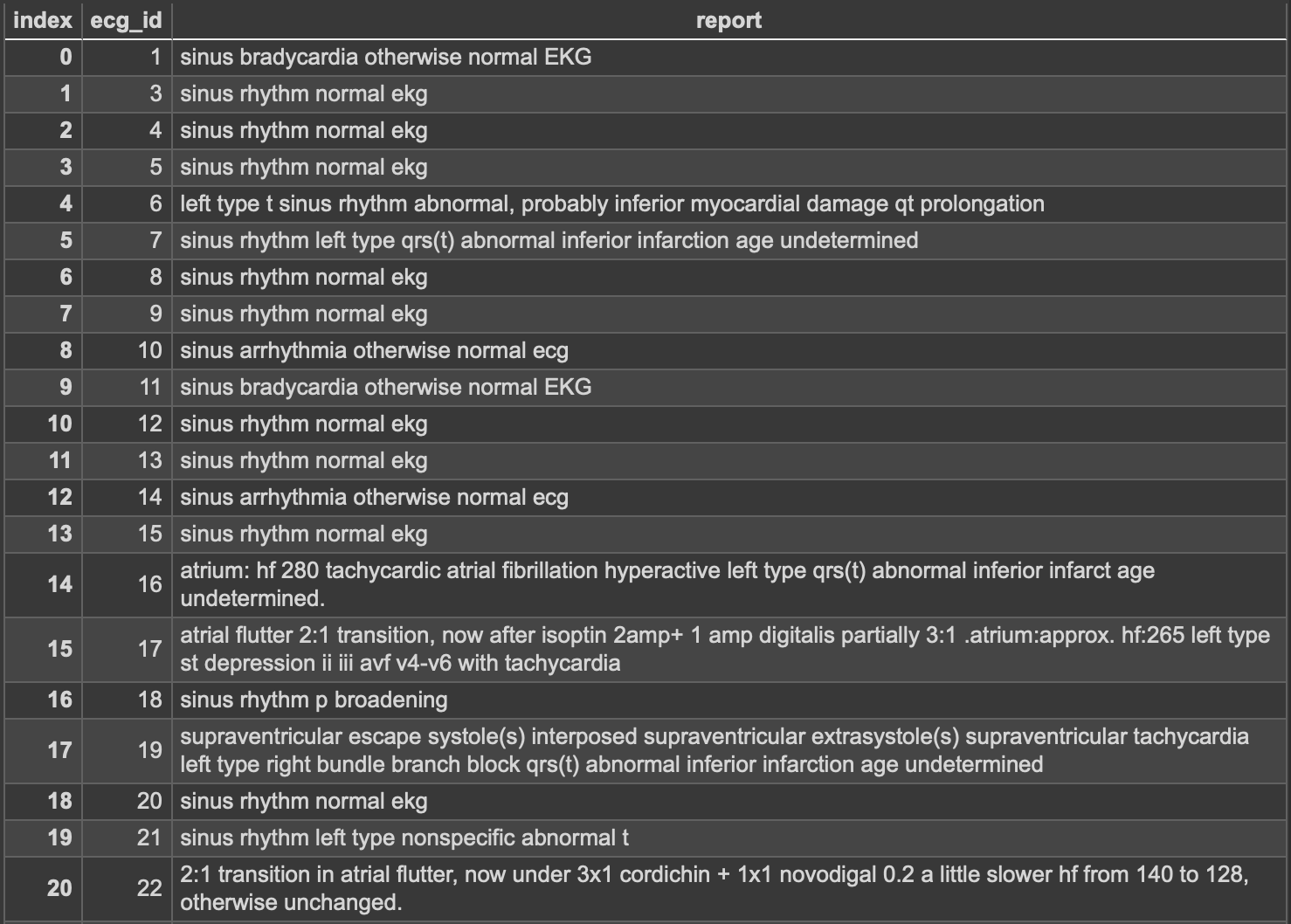}
\end{center}
\caption{Clinical Notes}
\label{datafig}
\end{figure}

\end{document}